# Similarity- based approach for outlier detection

[1]Amina Dik, [1]Khalid Jebari, [1,2]Abdelaziz Bouroumi and [1]Aziz Ettouhami

[1]LCS Laboratory, Faculty of Sciences, Mohammed V-Agdal University, UM5A

Rabat, Morocco

a.dik70@yahoo.fr, bouroumi@fsr.ac.ma, touhami@fsr.ac.ma

[2] LTI Information Treatment Laboratory, Ben M'sik Faculty of Sciences, Hassan II Mohammedia University (UH2M)

Casablanca, Morocco

a.bouroumi@gmail.com

**Abstract**

This paper presents a new approach for detecting outliers by introducing the notion of object's proximity. The main idea is that normal point has similar characteristics with several neighbors. So the point in not an outlier if it has a high degree of proximity and its neighbors are several. The performance of this approach is illustrated through real datasets.

## 1. Introduction

Placing a noise point in an existing cluster affects the quality of the results provided by clustering algorithms. Then clustering algorithms may generate misleading partitions arising outliers. Some methods have been proposed to detect outliers [1, 2] and a new concept of noise cluster was introduced [3, 4]. Unfortunately, these algorithms require some parameters which are not trivial to estimate.

An important part of the clustering tasks is a preliminary operation on the data that identifies outliers.

This paper presents an approach that deals with the outliers' problem by introducing the notion of object's degree of isolation and offers the possibility of eliminating or not such points.

## 2. Related work

An outlier is defined as an item that is considerably dissimilar from the remainder of the data [5]. In most cases, outliers are far away from all the other items without neighbors. Recently, some approaches have been proposed on outlier detection [6, 7] and the outliers themselves become the "focus" in outlier mining tasks [8].

These approaches can be classified into distribution-based and proximity-based approaches.

Distribution-based approaches, where outliers are defined based on the probability distribution [9, 10], develop statistical models. Items that have low probability to belong to the statistical model are declared as outliers [11].

The main idea in proximity-based methods is to consider outlier as an isolated point which is far away from the remaining data. This modeling may be performed in one of three ways. Especially, the three methods are distance-based, clustering-based or density-based approaches.

Distance-based approaches consider a point "$x$" as an outlier if there are less than "$M$" points within the distance "$d$" from "$x$". However, as the values of "$M$" and "$d$" are decided by the user, it is difficult to determine their values [6, 7].

Clustering-based approaches argue that outliers form very small clusters while normal objects belong to dense clusters [12, 13].

Others approaches are based on density. They compute the region's density in the data and consider items in low dense regions as outliers [14]. They consist to assign an outlying degree to each data point. This degree represents "how much" this data point is an outlier. The main difference between clustering-based approaches and density-based methods is that the first segment the points, whereas the second segment the space.

## 3. Proposed method

The proposed method to detect outliers is based on the notion of degree of proximity. This notion reflects the closeness of a point to other considered points. Here the meaning of closer is determined by the sum of the similarity of the point to each other. This degree of proximity can be considered as an opposite of outlier factor that characterizes the data points. More this degree is high the data point is not an outlier. The proposed approach can be considered as a hybrid approach between distance-based and density-based approaches.

The key idea is that normal point has more neighbors with which it has similar characteristics. So the point has a high degree of proximity when its neighbors are several.

The proposed method does not require any notion of clusters. It just inform if a data item is an outlier or not. The user should decide to delete or not such points.

Let $X=\{x_1, x_2,…, x_n\} \subset \Re^p$ be a finite set of n unlabeled feature vectors $x_i$, where $x_i \in \Re^p$ represents a sample pattern and $x_{ij}$ its $j^{th}$ feature. The problem is to detect outliers among these vectors. For this purpose, the proposed formula of proximity's degree of each vector is defined by:

$$D(x_i) = \left( \sum_{\substack{j=0 \\ j \neq i}}^{n} sim(x_i, x_j) \right) \quad (1)$$

Where:

$$Sim(x_i, x_k) = 1 - \frac{\|x_i - x_k\|_A^2}{p} \quad (2)$$

$Sim(x_i, x_k)$ is the similarity between the objects $x_i$ and $x_k$. A is the positive definite $pxp$ matrix defined by [15]

$$A_{jt} = \begin{cases} (r_j)^{-2}, & j = t \\ 0, & \text{otherwise} \end{cases} \quad (3)$$

The factor $r_j$ represents the difference between the upper and the lower limits of the attribute's values. It is defined by:

$$r_j = \max_{1 \leq i \leq n}\{x_{ij}\} - \min_{1 \leq i \leq n}\{x_{ij}\}, \ 1 \leq j \leq p \quad (4)$$

Denote $D^1_{min}$, $D^2_{min}$, $D^3_{min}$ and $D^4_{min}$ the four less measure of degree of proximity, and $D_{range} = \max_{1 \leq i \leq n}(D(x_i)) - \min_{1 \leq i \leq n}(D(x_i))$ the difference between the point which has the most degree of proximity and the less one.

Compute the follows values:

$D^1_{min} / D_{range}$, $D^2_{min} / D_{range}$, $D^3_{min} / D_{range}$ and $D^4_{min} / D_{range}$.

If the difference between $D^1_{min} / D_{range}$ is so low than the others values, the vector corresponding to $D^1_{min}$ is considered as outlier. If not, there are not outliers in the dataset.

We can also compute the "top M" outliers, M chosen by the user, within the (M + 2) less degree of proximity.

Our approach does not require the distance *d* that the user should define in the distance-based approach. Moreover, it does not require segmenting the space or the points.

Afterwards, we decide if these outliers vectors are eliminated or not and use clustering algorithm.

## 4. Numerical results and discussions

To evaluate the performance of our method, experiments are conducted on five real-world datasets that are available from the UCI Machine Learning Repository [16]: Wine, Breast Cancer, Spect Heart, Haberman's Survival and Breasst Tissu(see Table 1).

Wine Dataset is a result of a chemical analysis of wines from three different cultivars. There are 13 attributes and 178 samples from three classes corresponding to three different cultivars with respectively 59, 79, and 48 samples per variety.

Breast Cancer dataset is a 9-dimensional pattern classification problem with 699 samples from malignant (cancerous) class and benign (non-cancerous) class. The two classes contain respectively 458 and 241 points.

The third dataset describes diagnosing of cardiac Single Proton Emission Computed Tomography (SPECT) images. There are 22 attributes and 267 samples from two classes corresponding to normal and abnormal patients, with respectively 55 and 212 samples per category.

The fourth dataset is Haberman's Survival dataset that is the result of a measure of 306 cases on the survival of patients who had undergone surgery for breast cancer. It is a 3-dimensional pattern classification problem from two classes.

The last example is Breast Tissue recognition dataset that is the result of a measure of Breast Tissue by Electrical Impedance Spectroscopy. It is a 9-dimensional pattern classification problem with 106 samples from six classes.

Table.1 describes the type of data and gives information about attributes, size and number of classes.

Table 1: Datasets description

| Dataset | No. of Samples | No. of Attributes | No. of Classes |
|---|---|---|---|
| Wine | 178 | 13 | 3 |
| BCW | 699 | 9 | 2 |
| SPECT Heart | 267 | 22 | 2 |
| Haberman's Survival | 306 | 3 | 2 |
| BreastTissu | 106 | 9 | 6 |

At first, we search if there are outliers in the considered dataset. For this, we compute de density for the vectors and search the four less values. The results in table.2 show that there are outliers for wine, Haberman's Survival and Breast Tissu datasets.

For the case of wine dataset, $D^1_{min}/D_{range} = 3.93$ whereas $D^2_{min}/D_{range}$, $D^3_{min}/D_{range}$ and $D^4_{min}/D_{range}$ have a higher values (respectively 4.12, 4.14 and 4.15). For the Balance scale dataset, $D^1_{min}/D_{range} = 0.85$ whereas $D^2_{min}/D_{range}$, $D^3_{min}/D_{range}$ and $D^4_{min}/D_{range}$ have a same value 1.25 For the Haberman's Survival, $D^1_{min}/D_{range} = 0.86$ whereas $D^2_{min}/D_{range}$, $D^3_{min}/D_{range}$ and $D^4_{min}/D_{range}$ have respectively 1.04, 1.23 and 1.27

Table 3: Outliers Indices

| Dataset | Object 1 | Object 2 | Object 3 | Object 4 |
|---|---|---|---|---|
| Wine | 121 | 158 | 146 | 59 |
| Haberman's Survival | 258 | 26 | 204 | 225 |
| BreastTissu | 102 | 86 | 97 | 105 |

Table 2 - Results for outlier's detection.

| Dataset | $D^1_{min}$ | $D^2_{min}$ | $D^3_{min}$ | $D^4_{min}$ | $D_{range}$ |
|---|---|---|---|---|---|
| BCW | 165,73 | 166,20 | 172,03 | 174,66 | 352,07 |
| Wine | 109,67 | 113,80 | 114,88 | 115,60 | 27,9 |
| Heart | 51,52 | 53,12 | 56,15 | 57,04 | 79,4 |
| Haberman's Survival | 112,02 | 136,03 | 160,45 | 165,49 | 130,37 |
| BreastTissu | 27,00 | 54,91 | 55,99 | 57,53 | 60,04 |

Table 2 (after part) - Results for outlier's detection.

| Dataset | $D^1_{mi}/D_{range}$ | $D^2_{mi}/D_{range}$ | $D^3_{mi}/D_{range}$ | $D^4_{mi}/D_{range}$ | $D^1_{mi}/D_{range}$ |
|---|---|---|---|---|---|
| BCW | 0,47 | 0,472 | 0,48 | 0,49 | 0,47 |
| Wine | **3,93** | 4,07 | 4,11 | 4,14 | **3,93** |
| Heart | 0,65 | 0,67 | 0,71 | 0,72 | 0,65 |
| Haberman's Survival | **0.86** | 1.04 | 1.23 | 1.27 | **0.86** |
| BreastTissu | **0.44** | 0.91 | 0.93 | 0.95 | **0.44** |

The obtained results were also confirmed by the representation of the dataset. Indeed, Figure.1 shows that the datasets Wine, Haberman's Survival and BreastTissu contain outliers, whereas BCW and Heart dataset are without outliers as given by figure.2.

Table 3 – indices of outliers

| Dataset | object 1 (outlier) | object 2 | object 3 |
|---|---|---|---|
| Wine | 121 | 158 | 146 |
| Haberman's Survival | 258 | 26 | 204 |
| BreastTissu | 102 | 86 | 97 |

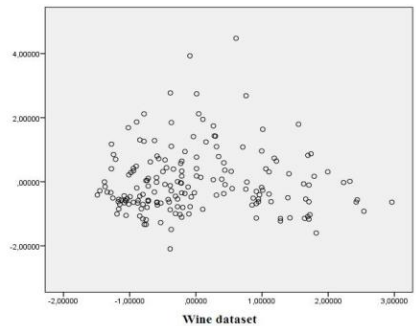

Wine dataset

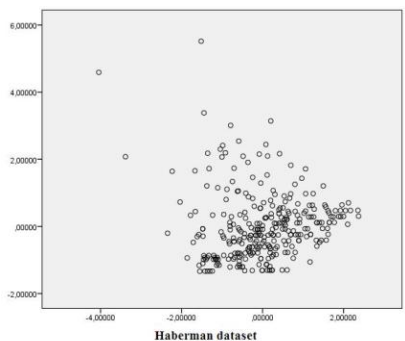

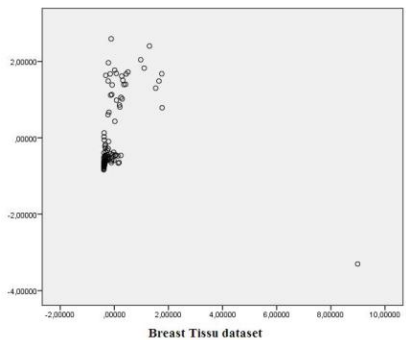

Figure 2- Representation of the dataset without outliers. (BCW and Heart)

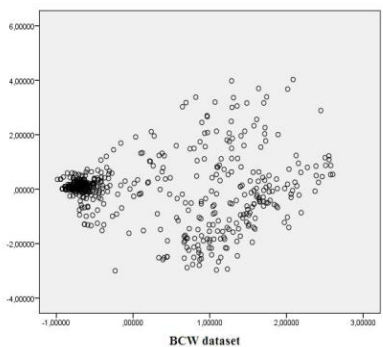

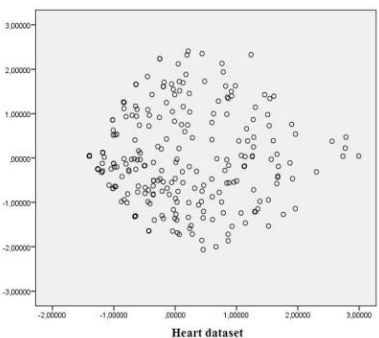

Several clustering algorithms are proposed in the literature. The most widely used clustering algorithm is FCM originally proposed by Bezdek [17]. Based on Fuzzy set theory, this algorithm partitions the considered objects such as similar objects are in the same cluster and dissimilar objects belong to different clusters.

To show the performance of FCM without outliers, we consider the M points outliers as centers for FCM and add a c centers. So we have a M + c clusters. We execute FCM with M+c clusters which have centers initialized with M points outliers and others c random points. In the end of processing, we verify if there are objects that belong to outliers cluster more than M.

## 4. Conclusions

In this paper presented an experimental study of a new method for detecting outliers. The proposed method is a hybrid approach between distance-based and density-based approaches. To improve the quality of clustering algorithms, our method can be used firstly to detect outliers and try to take a decision about them. We plan to improve and compare our technique in a future work.

**Amina Dik** PhD student in the Faculty of Sciences Rabat. Received his B.S. degree in Computer Science in 2002, and his M.S. in Computer Science, Multimedia and Telecommunications in 2008 from the Faculty of sciences, Rabat, Morocco.

**Khalid Jebari** Doctor in computer sciences. His research interests include Evolutionary Computation, Fuzzy Clustering, Design of new Meta-heuristic, Forecasting, Logistics and E-Learning.

**Abdelaziz Bouroumi** Professor and PhD. Advisor in the Faculty of Sciences, BenMsik, Casablanca, Morocco. His research include Soft Computing, E Learning, Fuzzy Clustering, He runs several projects in cooperation with other universities. He teaches several courses on operating system, neural network and computer architecture.

**Aziz Ettouhami** Professor and PhD. Advisor in the Faculty of Sciences, Mohamed V, Rabat, Morocco. His research include Soft Computing, Electronics, Telecommunications He runs several projects in cooperation with other universities. He is director of Conception and Systems Laboratory.